\documentclass[conference]{IEEEtran}
\IEEEoverridecommandlockouts
\usepackage{cite}
\usepackage{amsmath,amssymb,amsfonts}
\usepackage{algorithmic}
\usepackage{graphicx}
\usepackage{textcomp}
\usepackage{xcolor}
\usepackage{url}
\def\BibTeX{{\rm B\kern-.05em{\sc i\kern-.025em b}\kern-.08em
    T\kern-.1667em\lower.7ex\hbox{E}\kern-.125emX}}
\begin{document}

\title{Radio Frequency Signal based Human Silhouette Segmentation: A Sequential Diffusion Approach}

\author{
    \IEEEauthorblockN{
        Penghui Wen\IEEEauthorrefmark{2}\textsuperscript{,1}, 
        Kun Hu\IEEEauthorrefmark{2}\textsuperscript{,2,*}\thanks{* Corresponding author.}, 
        Dong Yuan\IEEEauthorrefmark{3}\textsuperscript{,3}, 
        Zhiyuan Ning\IEEEauthorrefmark{3}\textsuperscript{,4}
        Changyang Li\IEEEauthorrefmark{4}\textsuperscript{,5} and 
        Zhiyong Wang\IEEEauthorrefmark{2}\textsuperscript{,6}
    }
    \IEEEauthorblockA{
        \IEEEauthorrefmark{2}School of Computer Science, 
        The University of Sydney, Darlinton, NSW, Australia\\
    }    
    \IEEEauthorblockA{
        \IEEEauthorrefmark{3}School of Electrical and Computer Engineering, The University of Sydney, Darlinton, NSW, Australia\\
    }
    \IEEEauthorblockA{
        \IEEEauthorrefmark{4}Sydney Polytechnic Institute, Haymarket, NSW, Australia\\
    }
    \IEEEauthorblockA{
        \textsuperscript{1}pwen5103@uni.sydney.edu.au;
        \textsuperscript{2}kun.hu@sydney.edu.au;
        \textsuperscript{3}dong.yuan@sydney.edu.au;\\
        \textsuperscript{4}znin2645@uni.sydney.edu.au;
        \textsuperscript{5}chris@ruddergroup.com.au;
        \textsuperscript{6}zhiyong.wang@sydney.edu.au
    }
}


\maketitle

\begin{abstract}
Radio frequency (RF) signals have been proved to be flexible for human silhouette segmentation (HSS) under complex environments. Existing studies are mainly based on a one-shot approach, which lacks a coherent projection ability from the RF domain. Additionally, the spatio-temporal patterns have not been fully explored for human motion dynamics in HSS. 
Therefore, we propose a two-stage Sequential Diffusion Model (SDM) to progressively synthesize high-quality segmentation jointly with the considerations on motion dynamics. 
Cross-view transformation blocks are devised to guide the diffusion model in a multi-scale manner for comprehensively characterizing human related patterns in an individual frame such as directional projection from signal planes. 
Moreover, spatio-temporal blocks are devised to fine-tune the frame-level model to incorporate spatio-temporal contexts and motion dynamics, enhancing the consistency of the segmentation maps. 
Comprehensive experiments on a public benchmark - HIBER  demonstrate the state-of-the-art performance of our method with an IoU 0.732. 
Our code is available at \url{https://github.com/ph-w2000/SDM}.
\end{abstract}


\section{Introduction}

Human Silhouette Segmentation (HSS) aims to identify and separate human figures from the background in images. HSS has been used in a wide range of applications including human re-identification~\cite{guo2023multi}, fall detection~\cite{yu2017computer} and gait recognition~\cite{hu2019vision,hu2023multi}. The primary methodology involved the use of optical cameras for the segmentation of the human silhouette from RGB images~\cite{kirillov2020pointrend,kirillov2023segment,amit2021segdiff}. With the great success of deep learning methods across various domains~\cite{wen2023robust,tang2022otextsum,li2024towards,chen2023ccsd}, such techniques have been adopted for HSS as well using optical cameras, where they have demonstrated encouraging performance~\cite{xie2021segformer}. 
However, the inherent physical limitations result in significant performance degradation in capturing fine-grained human information, especially in low illumination and occlusion conditions. Moreover, optical cameras may give rise to privacy infringement issues~\cite{zhang2021widar3}. Therefore, researchers delved into privacy-preserving wireless sensors as potential solutions in such intricate environments.

\begin{figure}[h]
    \centering
    \includegraphics[width=\linewidth]{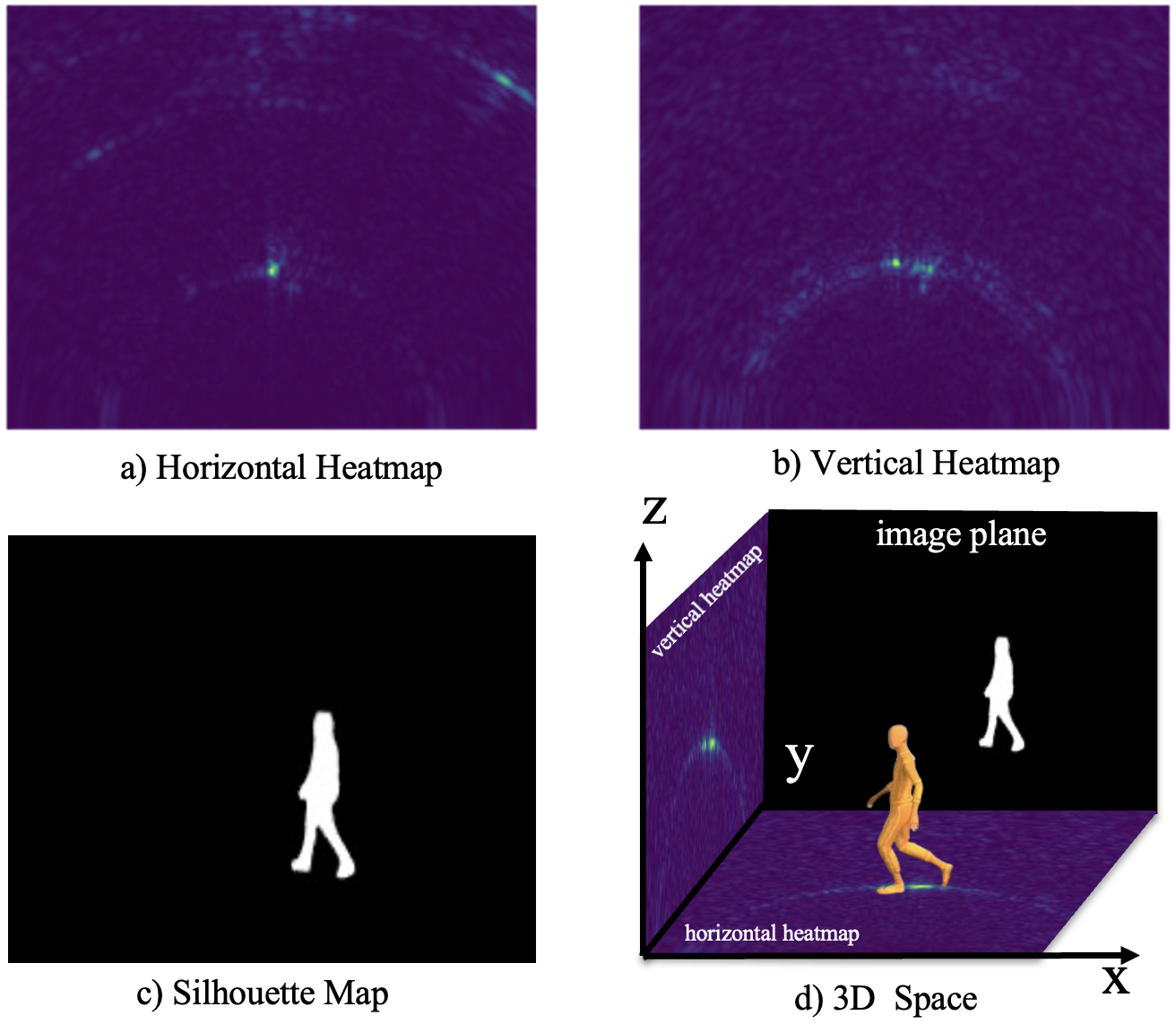}
    \caption{Human silhouette segmentation based on RF signals. Processed a) horizontal and b) vertical heatmaps converge to generate the c) silhouette map; and d) a 3D space represents an overall procedure of the RF-based segmentation task.}
    \label{fig:input_display}
\end{figure}

With the advancement of wireless technologies, wireless signals have found extensive applications in recognising human movement. For instance, in the domain of gesture recognition, a WiFi-based zero-effort cross-domain dataset was developed by~\cite{zhang2021widar3} for gesture recognition, while a mmWave data augmentation framework was devised by~\cite{li2022towards} for their gesture recognition system. In the field of pose estimation, the pioneer work proposed by~\cite{wang2019person} attempted to perceive human and estimate 2D human pose using WiFi signals. Subsequently, the use of WiFi signals was extended by~\cite{geng2022densepose} to achieve dense 3D human pose estimation through deep learning architectures. MobiRFPose~\cite{yu2023mobirfpose} was proposed to use radio frequency (RF) signal to estimate 3D human pose with a lightweight model and OCHID-Fi~\cite{zhang2023ochid} introduced the first RF-based 3D hand pose estimation method. More recently, RFMask~\cite{wu2022rfmask} framework has been introduced, which is the initial endeavor to generate human silhouette maps based on wireless RF signals.

As shown in Fig.~\ref{fig:input_display}, in this study, HSS is conducted using horizontal and vertical heatmaps processed from the collected RF signals. HSS characterizes human motion information including position, orientation and shape, and generates silhouette mask on the imaging plane orthogonal to the two signal heatmap planes. 
Existing methods like RFMask endeavors to generate silhouette maps in a single forward process, resulting in two major issues: 1) such one-shot way is hard to preserve a coherent projection of silhouette patterns from the two signal planes; and 2) the temporal patterns are not adequately explored, which are crucial for ensuring the consistency and precision regarding the human motions.

Therefore, we propose a two-stage deep learning framework for HSS, with a sequential diffusion model (SDM) for progressively synthesising high-quality HSS results. 
To condition on the diffusion process, multi-scale guidance is introduced to characterize foreground human patterns, which encodes the horizontal and vertical view heatmaps that are processed from RF signals. Then, cross-view transformation blocks are devised to inject the multi-scale patterns into the UNet of the diffusion pipeline such as directional projection from RF signals. 
Moreover, conventional diffusion methods typically focus on singleton image generation. To conduct a sequential HSS to involve continuous human motion dynamics, the two-stage learning strategy first optimize the silhouette diffusion pipeline at frame-level, and next a fine-tuning step is adopted for enhancing the consistency regarding the contexts at sequence-level with our proposed spatio-temporal blocks. 
Extensive experiments on a public benchmark - HIBER demonstrate the effectiveness of the proposed method, showcasing the state-of-the-art performance with an IoU 0.732.

The primary contributions of this study include:
\begin{itemize}
  \item The first diffusion approach for  human silhouette segmentation task from RF based multi-view heatmaps with a novel two-stage learning strategy.
  \item A cross-view transformation block (CTB) to guide the diffusion pipeline in a multi-scale manner to inject patterns such as directional projection from RF signals.
  \item A spatio-temporal block (STB), to fine-tune a singleton frame-based model to involve sequential context and motion dynamics.
\end{itemize}

\section{RELATED WORK}
Firstly, we introduce both optical camera-based and wireless sensor-based HSS. Subsequently, we delve into the evolution of diffusion models.

\subsection{Optical Cameras based HSS}

Over the past decade, optical based methods with cameras have shown encouraging outcomes in HSS. For example, Mask R-CNN~\cite{he2017mask} utilised a region-based convolution method to perform HSS and PointRend~\cite{kirillov2020pointrend} proposed a subdivision strategy for computing high-resolution silhouette maps efficiently. Segformer~\cite{xie2021segformer} eliminated the need for positional embedding and intricate decoder design used in conventional Transformer structure. Mask2Former~\cite{cheng2022masked} introduced masked attentions to limit attention within the foreground region of the predicted mask, leading to more efficient computing. Recently, the Segment Anything Model (SAM)~\cite{kirillov2023segment} demonstrated a significantly potent segmentation capability and propelled image segmentation into the era of foundation models. 

\subsection{Wireless Sensors based HSS}
Unlike optical cameras, wireless signals including LiDAR point clouds, WiFi signals, and radio frequency (RF) signals are minimally affected by complex environmental factors such as low illumination and occlusion. An efficient and scalable LiDAR-based online mapping method was studied~\cite{droeschel2018efficient}, and \cite{li2023memoryseg} introduced a latent memory mechanism to leverage temporal context information from sequential LiDAR inputs.
In contrast to LiDAR which offers high definition but comes with an expensive cost and susceptibility to weather, WiFi is cheaper and power efficient~\cite{hu2023muse}. The first study to detect individuals using pervasive WiFi signals was proposed in~\cite{wang2019person}. A 3D skeleton-based human pose estimation system was explored in~\cite{ren2022gopose} and Wi-Mesh~\cite{wang2022wi} leveraged WiFi signals to construct 3D human mesh.
RF signals are with broader bandwidth can provide more spatial information for more precise segmentation. RFMask~\cite{wu2022rfmask} performs HSS in an orthogonal plane using two signal planes in a single forward process. However, its one-shot generation process could miss fine-grained patterns and struggle to preserve coherent transformation from signal planes to image plane.

\subsection{Diffusion Methods}
Diffusion models~\cite{ho2020denoising, ho2022video, nichol2021improved,lu2024autoregressive}, have emerged as powerful generative models that are capable of synthesizing high-quality images. Following their success in visual content generation, these models have been adapted for segmentation tasks, such as DDP~\cite{ji2023ddp}, the aggregation \& decoupling framework~\cite{wang2024towards} and LD-ZNet\cite{PNVR_2023_ICCV}. Despite this progress, diffusion models still face limitations when it comes to segmenting wireless signals (in heatmaps) from multi-views. 
Therefore, our work aims to pioneer a novel diffusion-based method specifically tailored for HSS task using RF-based multi-view heatmaps. We also introduce a cross-view transformation block to guide the diffusion pipeline in a multi-scale manner for characterizing human related patterns and a spatio-temporal block to fine-tune the frame-level model to incorporate spatio-temporal contexts and motion dynamics.

\section{METHODOLOGY}

\begin{figure*}[h]
    \includegraphics[width=\linewidth]{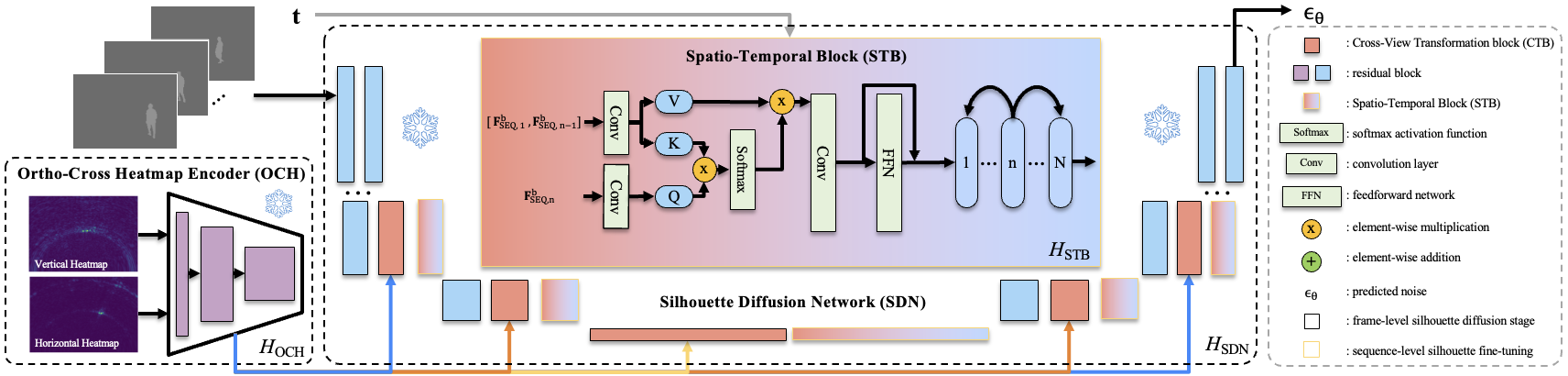}
    \caption{Illustration of the proposed architecture: sequential diffusion model (SDM). SDM is a UNet-based network composed of a ortho-cross encoder $H_{\text{OCH}}$ and a silhouette diffusion network $H_{\text{SDN}}$ with multiple cross-view transformation blocks (CTBs) and spatio-temporal blocks (STBs). The ortho-cross encoder $H_{\text{OCH}}$ encodes human motion patterns from the given paired horizontal heatmap and vertical heatmaps with residual blocks. To transfer rich multi-scale features extracted from $H_{\text{OCH}}$ to the $H_{\text{SDN}}$ , we propose to use cross-attention based cross-view transformation blocks (CTBs) that are embedded in different layers of $H_{\text{SDN}}$. During the \textit{Sequence-Level Silhouette Fine-Tuning} stage, we fix the weights of the components in the frame-level stage and insert an spatio-temporal block (STB) after every CTB block for fine-tuning, to formulate thespatio-temporal cohension.}
    \label{fig:architecture}
\end{figure*}

As illustrated in Fig.~\ref{fig:architecture}, a diffusion model $p_{\theta}(\mathbf{y}|\mathbf{c}_{\text{\text{hor}}},\mathbf{c}_{\text{ver}})$ is devised conditioning on a given horizontal heatmap $\mathbf{c}_{{\text{hor}}}$ and a vertical heatmap $\mathbf{c}_{{\text{ver}}}$. The desired output silhouette map $\mathbf{\hat{y}}$ is expected to adhere to the requirement of matching human motions information of the ground truth $\mathbf{y}$. In particular, the predicted $i^\text{th}$ pixel label $\hat{y}_i$ is expected to align with the corresponding ground truth $y_i$. The overall pipeline consists of two stages: \textit{frame-level silhouette diffusion} stage and \textit{sequence-level silhouette fine-tuning} stage. The \textit{frame-level silhouette diffusion} phase aims to perform silhouette segmentation based on individual frames, without taking into account the temporal relationships between consecutive frames. Meanwhile, the \textit{sequence-level silhouette fine-tuning} phase is devised to equip our model to address silhouette segmentation in a temporal context.

\subsection{Preliminary}
The foundation of our SDM approach relies on the denoising diffusion probabilistic model (DDPM)~\cite{ho2020denoising}. 
Denoting $\mathbf{y}_0 = \mathbf{y}$, DDPM involves creating a diffusion process that incrementally introduces noise to data sampled from the target distribution $\mathbf{y}_0 \sim q(\mathbf{y}_0)$, while the reverse denoising process is designed to learn an inverse mapping. The denoising process ultimately transforms the isotropic Gaussian noise $\mathbf{y}_{\text{t}} \sim \mathcal{N}(0, \mathbf{I})$ into the target data distribution over $t$ steps. 
In essence, this approach breaks down a challenging distribution-modelling task into a series of simple denoising problems. 
The forward diffusion path of DDPM forms a Markov chain characterized by the following conditional distribution:
\begin{eqnarray}
q(\mathbf{y}_{\text{t}} \mid \mathbf{y}_{\text{t-1}})=\mathcal{N}(\mathbf{y}_{\text{t}} ; \sqrt{1-\beta_{\text{t}}} \mathbf{y}_{\text{t-1}}, \beta_{\text{t}} \mathbf{I}),
\end{eqnarray}
where $t \in [1,T]$ and $\beta_1$,$\beta_2$,...,$\beta_{\text{t}}$ follow a fixed variance schedule with $\beta_{\text{t}} \in (0, 1)$. With the notation ${\alpha_{\text{t}}} = 1-{\beta_{\text{t}}}$ and $\hat{\alpha}_{\text{t}} = \prod_{s=0}^t \alpha_s$, we can derive samples from $q(\mathbf{y}_{\text{t}}|\mathbf{y}_0)$ in a closed form at any given timestep $t$: $\mathbf{y}_{\text{t}} = \sqrt{{\hat{\alpha}_{\text{t}}}}\mathbf{y}_0 + \sqrt{{1-\hat{\alpha}_{\text{t}}}}\epsilon$, where $\epsilon \sim \mathcal{N}(0, \mathbf{I})$~\cite{nichol2021improved}. A deep neural network can approximate the true posterior $q(\mathbf{y}_{\text{t-1}}|\mathbf{y}_{\text{t}})$ by predicting the mean and variance of $\mathbf{y}_{\text{t-1}}$. Specifically, two additional heatmaps $\mathbf{c}_{\text{hor}}$ and $\mathbf{c}_{\text{ver}}$ are used as conditions in this HSS task. The process can be achieved through the following parameterization:
\begin{eqnarray}
\begin{aligned}
p_\theta(\mathbf{y}_{\text{t-1}} | \mathbf{y}_{\text{t}}, \mathbf{c}_{\text{hor}}, \mathbf{c}_{\text{ver}})=&\mathcal{N}(\mathbf{y}_{\text{t-1}} ; \mu_\theta(\mathbf{y}_{\text{t}}, t, \mathbf{c}_{\text{hor}}, \mathbf{c}_{\text{ver}}),
\\
&\Sigma_\theta(\mathbf{y}_{\text{t}}, t, \mathbf{c}_{\text{hor}}, \mathbf{c}_{\text{ver}})),
\end{aligned}
\end{eqnarray}
where $\mu_\theta$ predicts the mean value and $\Sigma_\theta$ predicts the variance value of $\mathbf{y}_{\text{t-1}}$

\subsection{Frame-Level Silhouette Diffusion}

The \textit{frame-level silhouette diffusion} stage consists of an ortho-cross heatmap encoder $H_{\text{OCH}}$ and a UNet-based silhouette diffusion network (SDN) $H_{\text{SDN}}$. $H_{\text{OCH}}$ aims to formulate multi-scale features from the given horizontal and vertical heatmaps, while $H_{\text{SDN}}$ leverages the multi-scale features to generate silhouette segmentation maps. The details of the two components are discussed as follows.

\subsubsection{Ortho-Cross Heatmap Encoder}

The ortho-cross encoder $H_{\text{OCH}}$ encodes human motion patterns from the given paired horizontal heatmap $\mathbf{c}_{{\text{hor}}}$ and vertical heatmap $\mathbf{c}_{{\text{ver}}}$ with residual blocks. 
The encoded features extracted from the two heatmaps are subsequently combined and processed with a self-attention mechanism in pursuit of the pairwise consistency across the two planes. 
As different layers of the encoder observes heatmap associated patterns in different scales, 
leveraging their feature maps as multi-scale features is expected to obtain high-quality silhouette maps, particularly with regard to human limbs. 
Therefore, we 
construct a stacked feature representation $\mathbf{F}_\text{cond}=[\mathbf{F}^1_\text{cond}, \mathbf{F}^2_\text{cond}, ..., \mathbf{F}^\text{M}_\text{cond}]$, where $\text{M}$ represents the number of scales of $H_{\text{OCH}}$ output. 
To this end, the multi-scale features $\mathbf{F}_\text{cond}$ 
are ready to be used for guiding the silhouette diffusion network for generating segmentation masks.

\subsubsection{Discrete Segmentation Embedding}

Diffusion models are typically designed for continuous data, whilst silhouette segmentation map involves binary discrete labels. 
To address this, we adopt class embeddings, which employ a learnable embedding layer to map discrete labels $\mathbf{y}_{\text{0}}$ into a high-dimensional continuous space, normalized using a sigmoid function. The embedded features $\mathbf{y}^\prime_{\text{0}}$ will then be processed by our diffusion model. In the final stage of the diffusion model, a convolution layer is employed to convert the embeded features $\mathbf{y}^\prime_{\text{0}}$ into binary discrete labels $\mathbf{y}_{\text{0}}$ . For notation simplicity, we do not differentiate $\mathbf{y}_{\text{0}}$ and $\mathbf{y}^\prime_{\text{0}}$ in the following discussions.

\subsubsection{Silhouette Diffusion Network}

We predict the noise $\epsilon_\theta (\mathbf{y}_{\text{t}}, t, \mathbf{c}_{\text{hor}}, \mathbf{c}_{\text{ver}})$ using the silhouette diffusion network $H_{\text{SDN}}$. 
$t$ is the diffusion timestep,
and the noisy frame $\mathbf{y}_{\text{t}}$ is passed through $H_{\text{SDN}}$ to predict the noise added at step $t$. 
$H_{\text{SDN}}$ is a UNet-based network and additionally with our cross-view transformation blocks (CTBs). We insert a CTB between every two blocks of original UNet. 
CTB aims to guide this denoising process to inject patterns such as directional projection from RF signals based on cross-attentions, which inject the obtained $\mathbf{F}_\text{cond}$ into $H_{\text{SDN}}$ to embed multi-scale features in every layer of $H_{\text{SDN}}$. 

Formally, let $\mathbf{F}^\text{b}_{\text{frame}}$ be the input feature map regarding HSS in the $b^\text{th}$ UNet block of $H_{\text{SDN}}$. 
For the corresponding CTB, the keys $\mathbf{K}$ and values $\mathbf{V}$ are based on $\mathbf{F}_\text{cond}$, while queries $\mathbf{Q}$ are obtained from $\mathbf{F}^\text{b}_{\text{frame}}$. Formally, the computations of CTB can be formulated as:
\begin{eqnarray}
\begin{aligned}
&\mathbf{Q}=\textbf{W}^\text{b}_{\text{Q}} \circledast \mathbf{F}^\text{b}_{\text{frame}},
\mathbf{K}= \textbf{W}^\text{b}_{\text{K}} \circledast \mathbf{F}^\text{b}_\text{cond},
\mathbf{V}=\textbf{W}^\text{b}_{\text{V}} \circledast \mathbf{F}^\text{b}_\text{cond},
\\  
&\bar{\mathbf{F}}_{\text{frame}}^{\text{b}}=\mathbf{W}^{\text{b}}_{\text{CTB}} \text{softmax}(\frac{\mathbf{Q} \mathbf{K}^T}{\sqrt{d_\text{b}}}) \mathbf{V}+\mathbf{F}^\text{b}_{\text{frame}},
\end{aligned}
\end{eqnarray}
where $\textbf{W}^\text{b}_{\text{Q}}$, $\textbf{W}^\text{b}_{\text{K}}$ and $\textbf{W}^\text{b}_{\text{V}}$ are learnable $1\times 1$ convolution filters; and $\mathbf{W}^{\text{b}}_{\text{CTB}}$ is a learned weight matrix 
and $d_\text{b}$ is the vectorized dimension of these variables. 
Note that to compute 
$\bar{\mathbf{F}}_{\text{frame}}^{\text{b}}$, 
$\mathbf{Q}$, $\mathbf{K}$, and $\mathbf{V}$ are reshaped as matrices where each row indicates the channels of a pixel, and afterwards they are reshaped back to the original shape. 
The final output $\bar{\mathbf{F}}_{\text{frame}}^{\text{b}}$ is then processed by the $(b+1)^\text{th}$ UNet block in $H_{\text{SDN}}$.

\subsubsection{Frame-Level Optimization}
For the training of the denoising process, we generate a noisy sample $\mathbf{y}_{\text{t}} \sim q(\mathbf{y}_{\text{t}} |\mathbf{y}_0)$ by introducing a Gaussian noise $\epsilon$ to the ground truth $\mathbf{y}_0$. Subsequently, $\epsilon_\theta (\mathbf{y}_{\text{t}}, t, \mathbf{c}_{\text{hor}}, \mathbf{c}_{\text{ver}})$ is trained to forecast the introduced noise. A mean squared error (MSE) loss $\mathcal{L}_{\mathrm{MSE}}$ is adopted for the optimization:
\begin{eqnarray}
\mathcal{L}_{\mathrm{MSE}}=\mathbb{E}_{t \sim[1, T], \mathbf{y}_0 \sim q(\mathbf{y}_{\mathbf{0}}), \epsilon}\|\epsilon-\epsilon_\theta(\mathbf{y}_{\text{t}}, t, \mathbf{c}_{\text{hor}}, \mathbf{c}_{\text{ver}})\|^2.
\end{eqnarray}
The quality of the generated silhouette maps is further assessed using a cross-entropy loss $\mathcal{L}_{\mathrm{CE}}$, and a dice loss $\mathcal{L}_{\mathrm{DICE}}$.
\begin{eqnarray}
\mathcal{L}_{\mathrm{CE}}(\mathbf{\hat{y}}, \mathbf{y}) = - \sum_{i} y_i \log(\hat{y}_i),
\\
\mathcal{L}_{\mathrm{DICE}}(\mathbf{\hat{y}}, \mathbf{y}) = 1 - \frac{2 \sum_{i} y_i \hat{y}_i}{\sum_{i} y_i^2 + \sum_{i} \hat{y}_i^2}.
\end{eqnarray}
Note that 
we additionally adopt a loss function $\mathcal{L}_{\mathrm{VIB}}$, which aims to facilitate the learning regarding the variance $\Sigma_\theta$, following the existing practice~\cite{nichol2021improved}. 
To this end, the overall loss $\mathcal{L}$ is outlined as follows:
\begin{eqnarray}
\mathcal{L} = \lambda_1\mathcal{L}_{\mathrm{MSE}} + \lambda_2\mathcal{L}_{\mathrm{VIB}} + \lambda_3\mathcal{L}_{\mathrm{CE}} + \lambda_4\mathcal{L}_{\mathrm{DICE}},
\end{eqnarray}
where $\lambda_1$, $\lambda_2$, $\lambda_3$ and $\lambda_4$ are hyper-parameters to balance the weights of different loss terms.

\subsection{Sequence-Level Silhouette Fine-Tuning}

In the \textit{sequence-level silhouette fine-tuning} stage, our model is devised to accept sequential heatmaps 
as conditions to generate a sequence of silhouette maps with motion dynamics. 
In this stage, the weights of the components in the frame-level stage are fixed, 
and we additionally introduce spatio-temporal blocks (STBs) $H_{\text{STB}}$ based on cross-attentions. We insert an STB after every CTB block for fine-tuning, to formulate spatio-temporal cohesion regarding the motion dynamics. 



Due to the minimal differences between consecutive frames and the high computational complexity of an attention mechanism for all frames, we apply cross-attentions in a two-step manner. 
Let $\mathbf{F}^\text{b}_{\text{seq,n}}$ be the output feature map for the $n^\text{th}$ frame of the $b^\text{th}$ CTB block in $H_{\text{SDN}}$, where $\text{n} \in [2,\text{N}]$.
First, a cross-attention is adopted between a given frame $\mathbf{F}^\text{b}_{\text{seq,n}}$ and its two preceding frames, $\mathbf{F}^\text{b}_{\text{seq},1}$ and $\mathbf{F}^\text{b}_{\text{seq,n-1}}$. The leading frame $\mathbf{F}^\text{b}_{\text{seq},1}$ serves to define the target's position and orientation, while the former frame $\mathbf{F}^\text{b}_{\text{seq,n-1}}$ contributes to detailing the recent motion and shape.

Formally, the query embedding is formed by $\mathbf{F}^\text{b}_{\text{seq,n}}$, the key and value embeddings from $\mathbf{F}^\text{b}_{\text{seq},1}$ and $\mathbf{F}^\text{b}_{\text{seq,n-1}}$ . The cross-attention can be formulated as:
\begin{eqnarray}
\begin{aligned}
&\mathbf{Q}=\textbf{W}^\text{b}_{\text{Q}} \circledast \mathbf{F}^\text{b}_{\text{seq,n}}, \quad
\mathbf{K}=\textbf{W}^\text{b}_{\text{K}} \circledast ([\mathbf{F}^\text{b}_{\text{seq},1} \cdot \mathbf{F}^\text{b}_{\text{seq,n-1}}]), \\ 
&\mathbf{V}=\textbf{W}^\text{b}_{\text{V}} \circledast ([\mathbf{F}^\text{b}_{\text{seq},1} \cdot \mathbf{F}^\text{b}_{\text{seq,n-1}}]), \\
&\bar{\mathbf{F}}^{\text{b}}_{\text{seq,n}}=\operatorname{softmax}(\frac{\mathbf{Q} \mathbf{K}^T}{\sqrt{d}}) \mathbf{V},
\end{aligned}
\end{eqnarray}
where [$\cdot$] is a concatenation operation through the channel dimension and $d$ is the dimension of the associated variables. 
$\bar{\mathbf{F}}^{\text{b}}_{\text{seq,n}}$ is with a lower-dimension now, and a general sequential self-attention can be further adopted in the second step. 
The final output feature map is processed by the $(b+1)^\text{th}$ UNet block in $H_{\text{SDN}}$.


\section{Experiments}

\subsection{Dataset \& Implementation Details}

To demonstrate the effectiveness of the proposed method, we followed the same evaluation protocol and adopted HIBER dataset, the only publicly available dataset in RF-based HSS. It consists of \textit{WALK} and \textit{MULTI} subsets~\cite{wu2022rfmask}. In total, 152 group sequences in \textit{WALK} set  (walking with only one person in the scene) and 120 group sequences in \textit{MULTI} set  (walking with multiple persons in the scene) were collected. 
Each group sequence contains 590 silhouette frames with 590 horizontal and 590 vertical heatmaps. 
Considering a significant inter-group difference (between environment number 1 and others) in the distribution of the correlation between heatmap highlights and the ground truth human position, we eliminated these problematic data from our experiments. 
The evaluation was conducted both quantitatively and qualitatively, and intersection over union (IoU) was adopted as metrics. 

For the training of the first stage - \textit{frame-level silhouette diffusion}, an AdamW optimizer was adopted with a learning rate $2 \times 10^{-4}$. Training in this stage is conducted on one NVIDIA RTX3090, with a batch size 16. The multi-scale features $\mathbf{F}_\text{cond}$ consist of three scales, i.e., $5 \times 6$, $10 \times 12$ and $20 \times 25$. We set the values for $\lambda_1$, $\lambda_2$, $\lambda_3$ and $\lambda_4$ as 3, 1, 1 and 1.
For the second stage - \textit{sequence-level silhouette fine-tuning}, we employed two settings regarding the sequence length: 4 and 12, where the first setting was trained on one NVIDIA RTX3090 with a batch size 6 and the second setting was trained on one NVIDIA RTXA6000 with a batch size 5. The two stages were trained with 500 noising steps and a linear noise schedule. 

\subsection{Overall Performance}

Table~\ref{tab:quantity} lists quantitative comparisons between our proposed SDM and the state-of-the-art RFMask method. 
Note that SDM with sequence length 1 is with a setting only involving the first stage and processes singleton frames. 
The observation reveals that as the sequence length increases, the performance of SDM improves on both the \textit{WALK} and \textit{MULTI} sets, and SDM achieves the highest IoU score for both sets with a sequence length of 12. Compared to RFMask with the same sequence length, SDM outperforms RFMask on both sets. This suggests that SDM is capable of generating silhouette maps of superior quality.

\begin{table}[t]
\begin{center}
\caption{Quantitative comparison with RFMask.} \label{tab:quantity}
\begin{tabular}{lccc}
\hline
Model&\multicolumn{1}{l}{Seq. Length}&\multicolumn{1}{l}{WALK} & \multicolumn{1}{l}{MULTI}  \\ \hline
RFMask  & 4 & 0.681          & 0.682        \\
RFMask & 12 & 0.706          & 0.711        \\
SDM  & 1 &   0.689 &  0.686        \\
SDM & 4  &   0.714 & 0.702         \\ 
SDM  & 12 &  \textbf{0.732}& \textbf{0.716}            \\ \hline
\end{tabular}
\end{center}
\end{table}

\begin{figure}[h]
    \includegraphics[width=\linewidth]{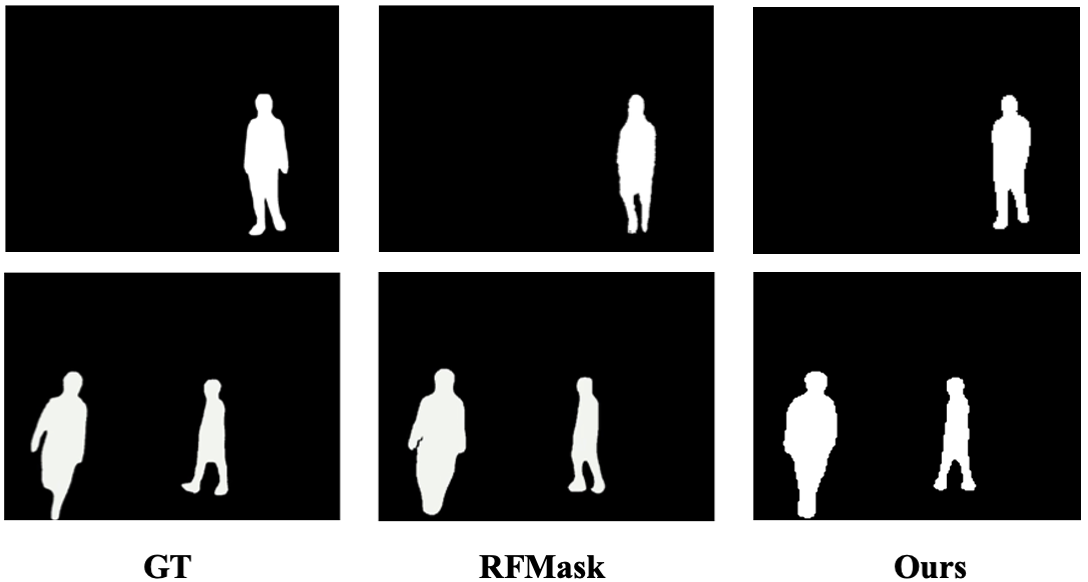}
    \caption{Qualitative comparison with RFMask.}
    \label{fig:quality}
\end{figure}

Fig.~\ref{fig:quality} provides a detailed visual comparison between our approach and RFMask on the \textit{WALK} and \textit{MULTI} sets of the HIBER dataset. The RFMask results are sourced from the authors' paper. 
It can be observed that SDM exhibits a better consistency preservation in human shape. For instance, regarding the example in the first row, the RFMask result resembles an alien-looking and produces implausible human leg volumes, whilst our result is more close to realistic human. In the second example, our model is capable of generating a slightly more realistic neck, whereas the RFMask result akins to a goitre human.

\subsection{Ablation Study}

\begin{figure}[h]
    \includegraphics[width=\linewidth]{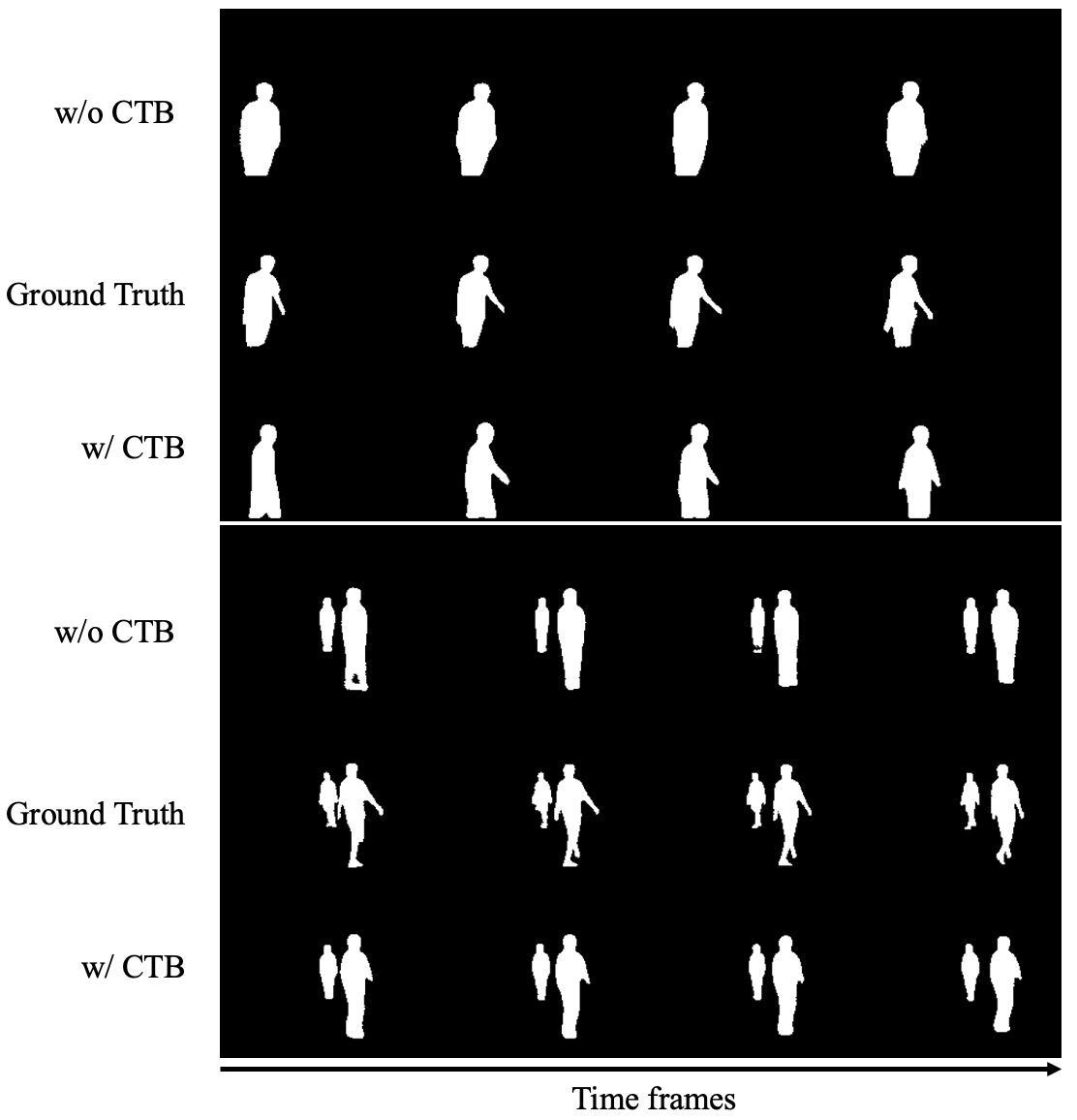}
    \caption{Ablation studies on CTB. The first example is from the \textit{WALK} set, while the second is from \textit{MULTI} set.}
    \label{fig:CTB_ablation}
\end{figure}

We conducted ablation studies to demonstrate the effectiveness of the proposed mechanisms. Table~\ref{tab:ablation_study} illustrates the impact of CTB and STB modules. 
For the base scenario, neither CTB units nor STB is used. This setup simply concatenates multi-scale features $\mathbf{F}_\text{cond}^\text{b}$ with $\mathbf{F}^\text{b}_{\text{frame}}$ and produces singleton silhouette frames. 
We integrate CTB into the base, denoted as base + CTB. In comparison with the base, base + CTB exhibits improvements in both sets regarding IoU. 
As shown in Fig.~\ref{fig:CTB_ablation}, base without CTB leads to inaccurate inter-plane projection, causing a frontal display instead of the intended side display. Conversely, when CTB is employed, the outcomes clearly demonstrate the rightward movement of the target. Therefore, the results indicate CTB is capable of modelling the intricate directional information from signal planes to silhouette plane.

\begin{figure}[h]
    \includegraphics[width=\linewidth]{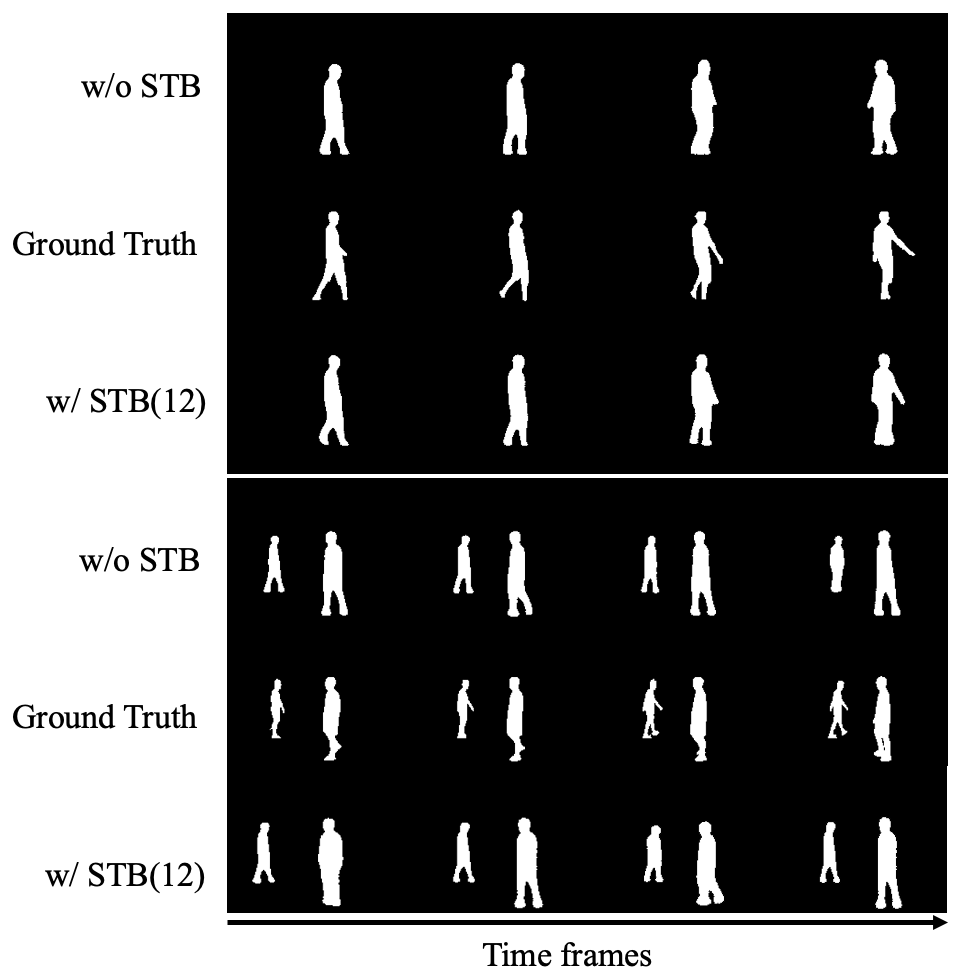}
    \caption{Ablation studies on STB. The first example is from the \textit{WALK} set, while the second is from \textit{MULTI} set.}
    \label{fig:consistency_problem}
\end{figure}

Next, we integrate our STB into base + CTB, denoted as base + CTB + STB, which takes 12 sequence frames as input. The results indicate an enhancement in both sets quantitatively. As shown in Fig.~\ref{fig:consistency_problem}, consistency issue occurs for base + CTB that the fourth frames are erroneously generated with an incorrect direction, while employing base + CTB + STB maintains consistent directional information. Therefore, our STB attention module can effectively enhance the consistency of sequential context and motion dynamics to improve the quality of silhouette maps.

\begin{table}[t]
\begin{center}
\caption{Ablation studies on the proposed mechanisms.} \label{tab:ablation_study}
\begin{tabular}{lccc}
\hline
Methods     & \multicolumn{1}{l}{Seq. length}  & \multicolumn{1}{l}{WALK}  & \multicolumn{1}{l}{MULTI}  \\ \hline
Base   & 1  & 0.669     & 0.671   \\
Base + CTB  & 1    &  0.689  & 0.686    \\ 
Base + CTB + STB & 12   & \textbf{0.732} & \textbf{0.716}   \\ \hline
\end{tabular}
\end{center}
\end{table}

\section{Conclusion}

We present a two-stage diffusion-based HSS approach - SDM using RF signals.  We introduced CTB and STB modules to formulate frame-level and sequence-level patterns. These allow SDM progressively synthesizing high-quality segmentation maps considering motion dynamics. 
The main limitation of SDM is the coarse segmentation of each person's limbs in scenes with multiple persons. We are considering the use of a region proposal network to treat each person individually as a potential solution in the future.
Additionally, addressing issues such as rough body edges and missing arms is essential to achieve more realistic human generation.

\section*{Acknowledgement}
This study was partially supported by Australian Research Council (ARC) grant DP210102674.

\bibliographystyle{IEEEbib}
\bibliography{ICME2024/icme2023template}

\end{document}